\documentclass[10pt,journal,compsoc]{IEEEtran}

\usepackage{graphicx}
\usepackage{amsmath}
\usepackage{amssymb}
\usepackage{xspace}
\usepackage{pifont}
\usepackage{hyperref}

\usepackage{tcolorbox}
\usepackage{microtype}

\usepackage{mathtools}
\usepackage{rotating}
\usepackage{siunitx}
\usepackage[capitalize,noabbrev]{cleveref}
\usepackage[textsize=tiny]{todonotes}
\usepackage{placeins} 

\usepackage{multirow}
\usepackage{amsfonts}
\usepackage{amsthm}
\usepackage{mathrsfs}

\usepackage[table]{xcolor}
\usepackage{textcomp}
\usepackage{manyfoot}
\usepackage{booktabs}
\usepackage{algorithm}
\usepackage{algorithmicx}
\usepackage{algpseudocode}
\usepackage{listings}
\usepackage[numbers,sort&compress]{natbib}
\usepackage{subcaption}
\usepackage{caption}
\usepackage{stfloats}


\newcommand{\cmark}{\ding{51}} 
\newcommand{\xmark}{\ding{55}} 
\newcommand{\foobar}{\includegraphics[width=.9em]{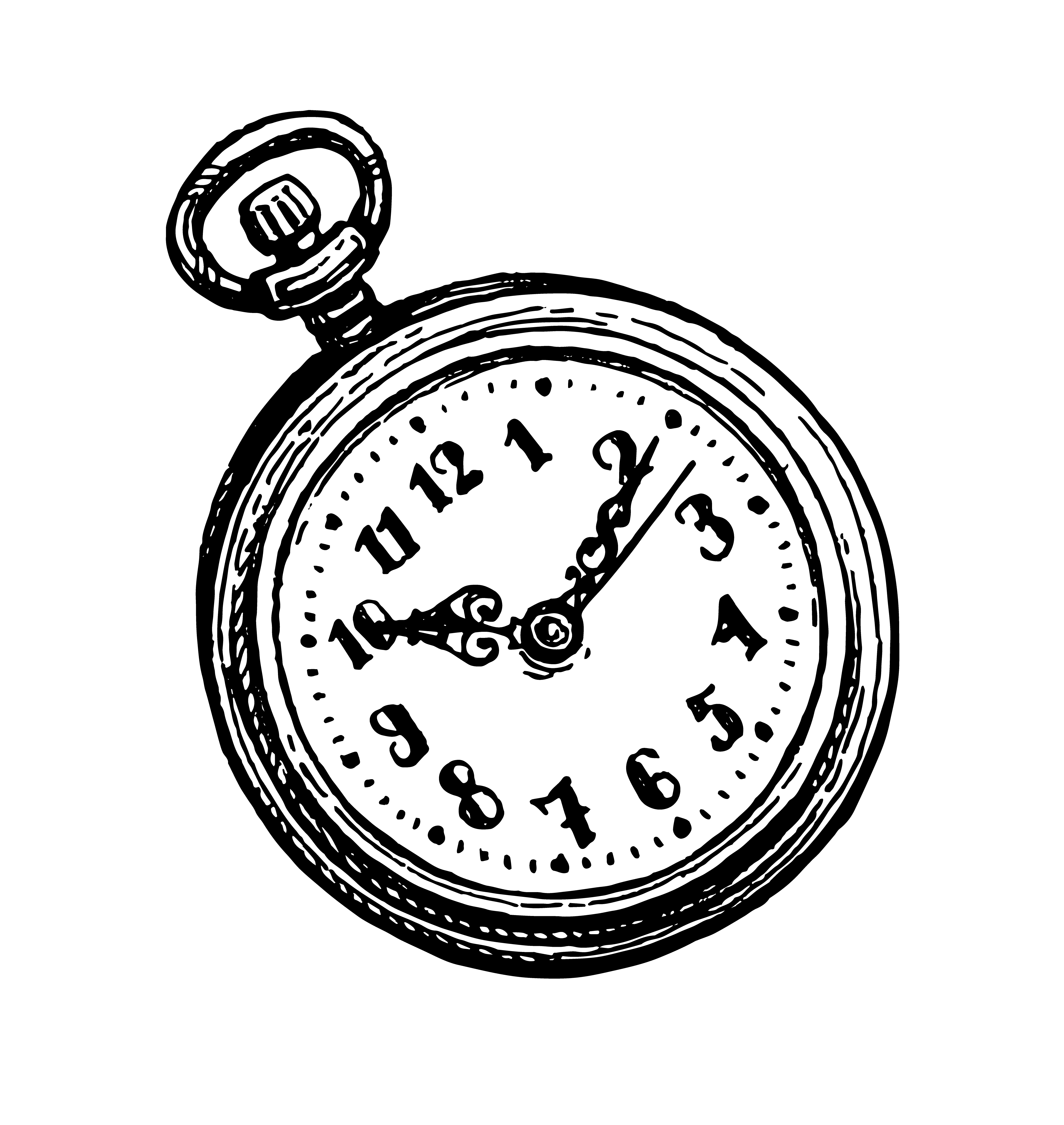}}
\newcommand{\chrono}{\emph{Chrono}\xspace}
\newcommand{\chronoBlip}{\emph{Chrono-BLIP}\xspace}
\newcommand{\chronoGPT}{\emph{Chrono-GPT}\xspace}
\newcommand{\chronoQwen}{\emph{Chrono-Qwen}\xspace}
\newcommand{\chronoWatch}{\emph{Chrono}\foobar\xspace}
\newcommand{\chronoQwenWatch}{\emph{Chrono-Qwen}\foobar\xspace}
\newcommand{\chronoGPTWatch}{\emph{Chrono-GPT}\foobar\xspace}
\newcommand{\chronoBlipWatch}{\emph{Chrono-BLIP}\foobar\xspace}
\definecolor{lightgray}{gray}{0.9}
\newcommand{\rowbg}{\cellcolor{lightgray}}

\newcommand{\myparagraph}[1]{\noindent\textbf{#1.}\xspace}

\begin{document}

\title{Chrono: A Simple Blueprint for Representing Time in Multimodal Large Language Models}

\author{Hector~G.~Rodriguez*, Boris~Meinardus*, Anil~Batra, Anna~Rohrbach, Marcus~Rohrbach}

\markboth{Transactions on Pattern Analysis and Machine Intelligence}{Rodriguez \MakeLowercase{\textit{et al.}}: Chrono Blueprint}

\maketitle

\begin{abstract}
The recent success of Large Language Models (LLMs) has prompted the extension to the multimodal domain, developing image-text Multimodal LLMs (MLLMs) and then video-text models.
In this work, we investigate the challenge of contextual and temporal comprehension in video-language models by exploring the task of \emph{temporal localization} in videos.
To address this problem, prior works have developed complex task-specific architectures, novel modules to embed time into MLLMs, or leveraged additional input signals such as video transcripts to best encode contextual and temporal information.
We find that most of these efforts are surpassed by a much simpler design.
We introduce \chronoWatch, a universal sequence blueprint that can be applied to any image-text pretrained MLLM.
In extensive experiments spanning different MLLM architectures and sizes, finetuning and zero-shot settings, we demonstrate new state-of-the-art results in moment retrieval on the widely used benchmarks Charades-STA, QVHighlights, and ActivityNet Captions, as well as in grounded video question answering on NExT-GQA.
\footnote{The code is available at \href{https://github.com/sudo-Boris/mr-Blip}{https://github.com/sudo-Boris/mr-Blip}.}

\end{abstract}


\section{Introduction}
\label{sec:intro}

The recent success of pretrained large language models (LLMs)~\cite{brown2020gpt3, zhang2022opt, chung2022t5} has inspired the development of generative image-text pretrained multimodal large language models (MLLMs)~\cite{alayrac2022flamingo, li2023blip2, huang2023kosmos1} that can comprehend vision and language modalities jointly.
However, due to higher computational and annotation costs, large-scale pretraining on video data is more demanding.
To circumvent the issue, recent studies leverage image-text pretrained models for image-to-video transfer learning~\cite{yu2023EILEV, yu2023SeViLA, lei2021CLIPBERT, luo2021clip4clip, fang2021clip2video, ju2022prompting}.
Such models offer promising results in the direction of video-text retrieval~\cite{luo2021clip4clip}, video captioning~\cite{xu2023mplug}, or multiple choice video question answering~\cite{yu2023SeViLA}. 
Yet, those tasks don't require precise temporal understanding, whereas the task of \emph{moment retrieval} (MR) requires the precise temporal localization of all moments associated with an open-ended natural-language query in an untrimmed video.
This ability has not been extensively explored for such models.

\begin{figure}[htbp]
  \centering
  \includegraphics[width=0.9\linewidth]{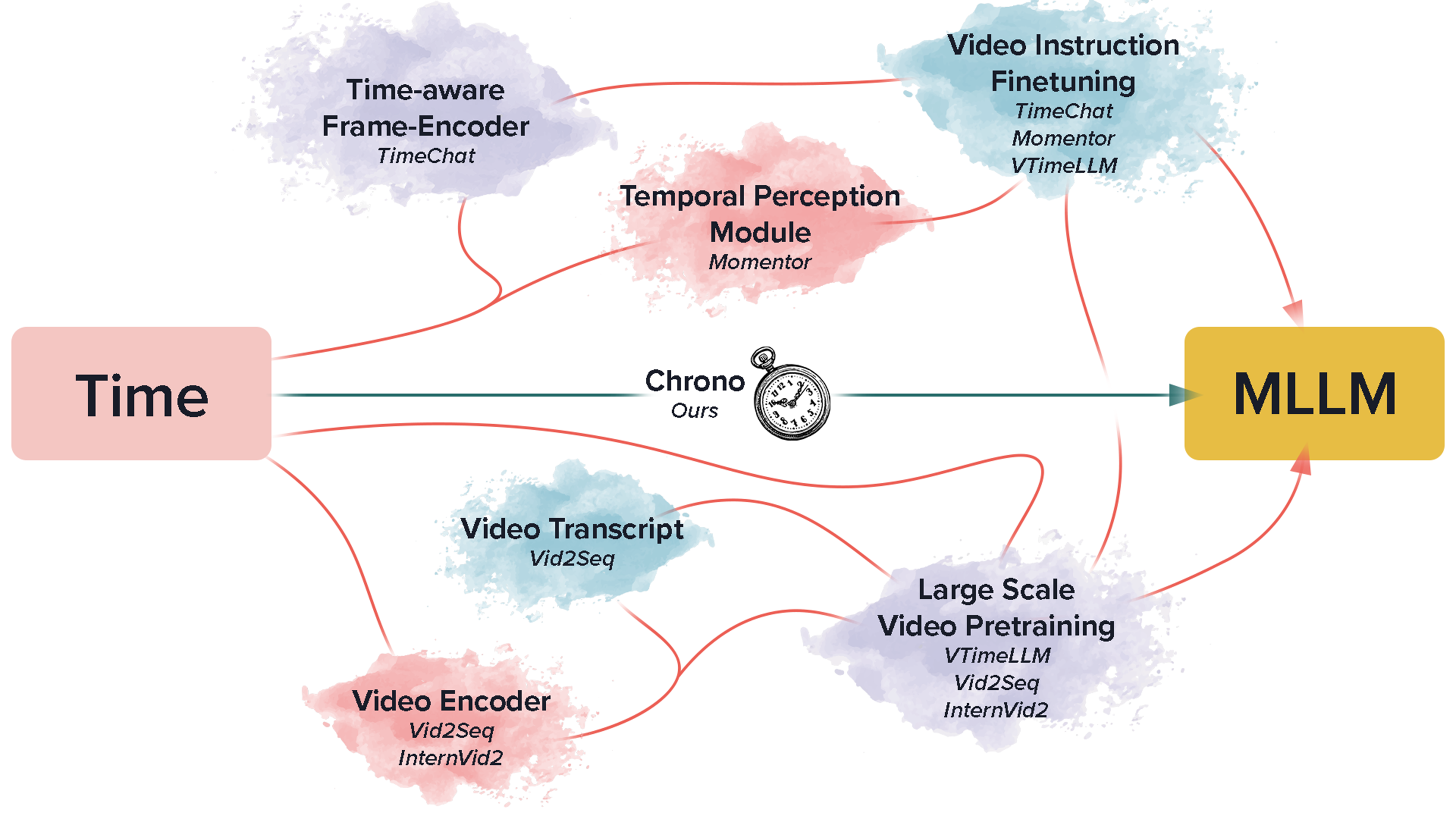}
  \caption{
      \textbf{How can we endow an MLLM with a sense of time?} In contrast to prior work, which leverages separate modules and diverse pretraining strategies, \chronoWatch natively allows the MLLM to understand the language of time and associate it with the corresponding segments in a video. 
  }
  \label{fig:teaser}
\end{figure}

The moment retrieval task has multiple valuable applications, such as video search or video indexing.
Additionally, we highlight temporal grounding as a test bed for temporal understanding of video-language models.
More concretely, it requires understanding, discrimination, and temporal localization of multiple events in potentially minutes-long videos given an open-ended natural-language query.
In the context of MLLMs, it is an open question of how to best model this task as a sequence-to-sequence prediction task.
Specifically, how to encode time best so the model can reason about it and predict start and end timestamps correctly.

Traditionally, prior works generally approach the challenge of moment retrieval by leveraging video features~\cite{feichtenhofer2019slowfast} or CLIP~\cite{radford2021CLIP} to train a complex, task-specific feature fusion module, which 
 ultimately predicts a fixed set of candidate windows.
As shown in \cref{fig:teaser}, recent approaches employ MLLMs for temporally grounded video-language tasks, one of the most relevant being moment retrieval.
However, they often require large-scale instruction tuning datasets~\cite{Ren2023TimeChatAT, qian2024momentor, yang2023vid2seq}, complex multi-stage training~\cite{huang2024vtimellm}, specialized architectures~\cite{qian2024momentor, Ren2023TimeChatAT}, or further input signal, such as video transcripts~\cite{yang2023vid2seq}.

In contrast, we develop \chronoWatch, a novel multimodal input sequence blueprint that enables image-text pretrained MLLMs to better understand time and achieve higher quality video moment retrieval.
\chronoWatch interleaves language timestamps with video frames, and appends duration information.
By using the native language input space of the model, it avoids adding special tokens or any architectural modification.
Extensive ablations demonstrate that this simple approach outperforms all prior methods, which add significantly more complexity, as shown in \cref{fig:teaser}.
We show that this input design enables the best temporal grounding abilities both when finetuning image-text pretrained MLLMs or when using larger MLLMs in a zero-shot manner.
\chronoWatch achieves state-of-the-art (SOTA) results across the widely used benchmarks Charades-STA~\cite{gao2017charadesSTA}, QVHighlights~\cite{lei2021qvhighlights}, and ActivityNet Captions~\cite{krishna2017ActivityNetCaptions}.
We discuss the relevance of deliberate design choices by conducting extensive ablation studies on MLLMs with different architectures, BLIP-2~\cite{li2023blip2}, Qwen2.5-VL \cite{Qwen2.5-VL} and GPT-4o~\cite{GPT4o}.
Besides the video moment retrieval datasets, we also consider NExT-GQA~\cite{Xiao2023NExT-QGA}, a dataset for grounded video QA (GVQA).
The task there is to localize temporal segment(s) relevant to the question's answer.
Here, we assess \chrono's ability to localize such video evidence without being trained for this task, as well as to generate an answer.
Again, we show constitent improvements over prior approaches which are significantly more complex.

The main contributions are as follows:
\textit{(i)} We leverage image-text pretrained MLLMs to approach moment retrieval by casting it as an open-ended sequence-to-sequence problem. 
\textit{(ii)} To enhance the temporal understanding of events in input videos, we design a novel multimodal input sequence and introduce \chronoWatch, a simple and universal blueprint for representing time in MLLMs.
\textit{(iii)} \chrono models, derived from BLIP-2, GPT-4o and Qwen2.5-VL, improve the state-of-the-art on the widely used moment retrieval benchmarks~\cite{gao2017charadesSTA, lei2021qvhighlights, krishna2017ActivityNetCaptions} and achieves a new state-of-the-art on grounded VQA~\cite{Xiao2023NExT-QGA}.
\textit{(iv)} Extensive experiments and ablations demonstrate the effectiveness of \chrono and its design choices, highlighting that deliberate exploration of simple methods can outperform complex, potentially over-engineered methods. 
\textit{(v)} We make the code and models publicly available.

Since the release of our early arXiv preprint \citep{meinardus2024surprising}, \chrono has been adopted by several works to improve temporal understanding in moment retrieval, question answering, and captioning.
More details are provided in \cref{sec:RelatedWork}.
An early version of this work appeared at the ICCV 2025 workshop on Multimodal Representation and Retrieval \citep{Meinardus_2025_ICCV}.

\section{Related Work}
\label{sec:RelatedWork}

\myparagraph{Image-to-Video Transfer Learning}
The limited public availability of high-quality large-scale video-language datasets, together with the immense compute requirements of training using videos, pose a significant challenge for large-scale video-language pretraining. The idea of leveraging image-language pretrained models for image-to-video transfer learning by utilizing a limited number of video frames to enhance learning efficiency has proven effective as evidenced by many recent publications~\cite{yu2023EILEV, yu2023SeViLA, cao2022locvtp, fang2021clip2video, Fang2022TransferimImageCLIP, ju2022prompting, lei2021CLIPBERT, luo2021clip4clip, ma2022xclip, lei2022revealing, xue2023clipvip, wang2022objectaware}.

In particular, \cite{yu2023SeViLA} makes use of the contextual understanding of MLLMs to perform video QA tasks.
The authors train a BLIP-2 model to individually select 4 question-relevant frames and, using those, fine-tune another BLIP-2 to answer the question.

We also leverage the pretrained BLIP-2 model and finetune it on the downstream task of moment retrieval, achieving significant localization improvements over \citet{yu2023SeViLA}.
Furthermore, we show that strong image-text models like GPT-4o~\cite{openai2024gpt4technicalreport}, which cannot perform moment retrieval natively, achieve competitive results using the \chrono blueprint.

\myparagraph{Moment Retrieval Models}
Moment Retrieval (MR) tasks requires to analyze a video and find the relevant clip given an open-ended natural language query.
Conventional approaches fall into either proposal-based or proposal-free methods.
Proposal-based methods learn to identify moment candidates given predefined proposals, e.g., sliding windows~\cite{hendricks2017localizing, gao2017charadesSTA} and temporal anchors~\cite{chen2018temporalgrounding, wang2019temporally}, in a first stage.
In a second stage, these candidates are further refined to better match the text query.
Regression-based methods~\cite{lei2021qvhighlights, yan2023unloc, liu2022umt, mun2020localglobal, zeng2020dense} are common among proposal-free approaches. 
These directly predict the temporal boundaries of a relevant moment, without needing a first stage to generate initial proposals.
Nevertheless, proposal-free methods still predict a fixed number of candidate windows alongside a confidence score, based on which the predictions are sorted. 
Using a fixed number of proposals has some downsides, like not generalizing to videos with a higher number of actions than the data used for tuning the number of proposals hyperparameter or wasted computation on duplicate proposals.

\myparagraph{MLLM-based temporal grounding}
With the advent of multimodal large language models (MLLMs), several works have approached MR by using multimodal LLMs, leveraging their visual capabilities and extensive world knowledge.
However, powerful MLLMs still struggle to understand and express time precisely, which impedes their direct use for MR.
Therefore, \citet{Ren2023TimeChatAT, huang2024vtimellm, qian2024momentor, huang2025lita} focus on \emph{how to represent time} inside MLLMs.
\citet{Ren2023TimeChatAT} introduce time-sensitive frame embeddings by incorporating temporal information through Q-Former, into their model, TimeChat.
\citet{huang2024vtimellm} train VTimeLLM three stages, including extensive multiple-event video-language pretraining and high-quality video-instruction tuning.
\citet{qian2024momentor} present Momentor, which uses a dedicated Temporal Perception Module.
\citet{huang2025lita} explore the use of special temporal tokens for LITA. 
These works, require large-scale instruction-tuning datasets (TimeChat~\cite{Ren2023TimeChatAT}, Momentor~\cite{qian2024momentor}), complex multi-stage training (VTimeLLM~\cite{huang2024vtimellm},LITA~\citet{huang2025lita}), or specialized architectures (Momentor~\cite{qian2024momentor},LITA~\citet{huang2025lita}).
Distinct from that, this work shows that careful design choices in representing temporal information can enable moment retrieval using off the shelf MLLMs without any architectural modifications. 
Furthermore, when combined with direct finetuning on downstream tasks, the \chronoWatch blueprint introduced in this work can achieve superior performance without the need for expensive pretraining or architectural complexity.
Through systematic ablations, we demonstrate that simple design choices - such as using absolute integer timestamps - can outperform more complex approaches that introduce special tokens or dedicated temporal modules.
Alternative, recent work has shown that timestamping beyond language, i.e. frames directly in visual input space, is useful when using models with more powerful visual capabilities \citet{numberit2025}.

\myparagraph{Influence of the \chronoWatch{} blueprint on subsequent work}
Inspired by the early arXiv preprint version of this work \citep{meinardus2024surprising}, several recent methods employ the \chronoWatch{} blueprint to achieve strong performance on moment retrieval tasks.
\citet{lu2024llavamrlargelanguageandvisionassistant} extend the LLaVA vision-language model for moment retrieval by using \chronoWatch{} blueprint,
in addition to frame selection and token compression to process longer videos.
Similarly, \citet{unitime2025} employ the \chronoWatch{} blueprint, together with an adaptive per frame token rate and hierarchical segment prediction at inference, to better handle the moment retrieval task in long videos. 
\citet{zeng2025tempo} further develop the \chronoWatch{} blueprint to employ a fixed number of tokens per timestamp,
and use RL to teach the model to perform moment retrieval in a more flexible manner.

Besides moment retrieval, time representation is also relevant for multiple video tasks. 
The \chronoWatch{} blueprint has also been used to improve temporal understanding in a variety of video tasks.
\citet{dtos2025} employ \chronoWatch{} to obtain better temporal representation, which proves crucial in the Referring video Object Segmentation task. 

Additionally, \citet{videollama3} adopt a similar timestamping scheme, demonstrating improved performance across diverse video QA settings, including general and long-video understanding, temporal reasoning, and temporal grounding. 

The impact of \chronoWatch{} goes beyond influencing the way time is instilled in video inputs, as its SOTA performance as a moment retrieval model is used by \citet{zeng2025distimedistributionbasedtimerepresentation} to pseudo-label the segments corresponding to video captions.
The resulting model demonstrates enhanced dense video captioning and video question answering, in addition to moment retrieval.

\section{\chronoWatch: A Blueprint for Time Representation in MLLMs}
\label{sec:Method}
\begin{figure*}[tb]
  \centering
  \includegraphics[width=\textwidth]{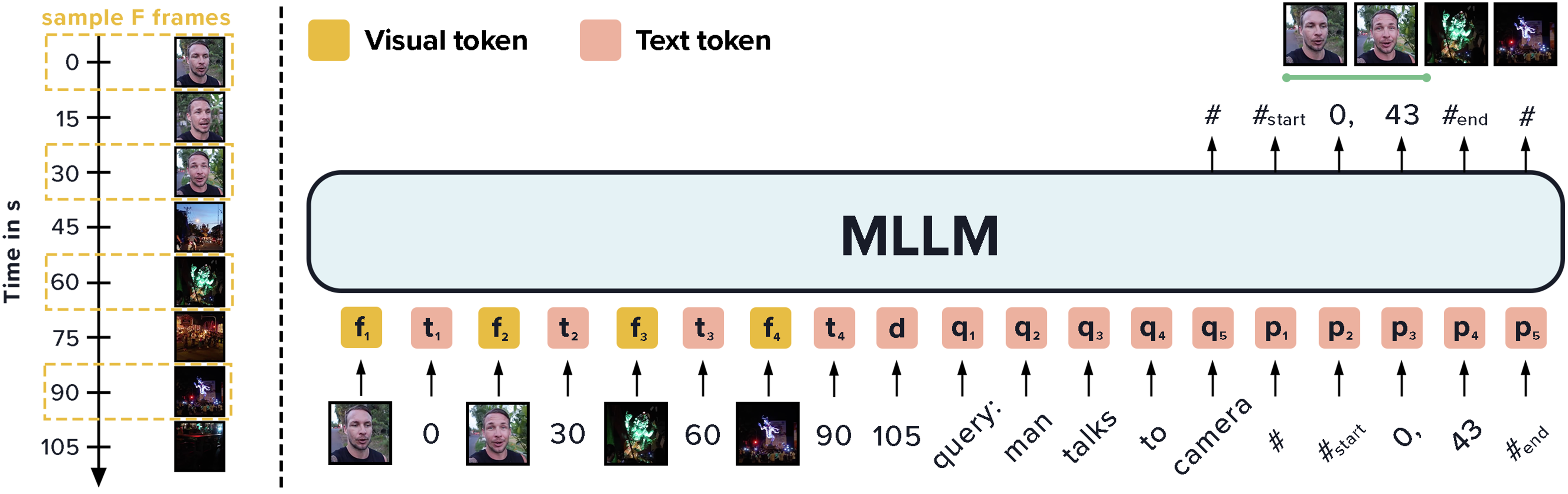}
  \caption{
    \textbf{\chronoWatch model overview.}
    The \chronoWatch{} blueprint is model-agnostic and can adapt an image-text pretrained MLLM even in a zero-shot (no finetuning) setting.
    We construct the input for the MLLM by interleaving the frame embeddings and the timestamps of each sampled frame, followed by the video duration, the moment retrieval query, and a task prompt (the latter being not visualized.)
    The MLLM outputs a sequence of potentially multiple retrieved moments by predicting the global BOS and EOS tokens, denoted by $\#$, the start of window and end of window tokens denoted as $\#_{start}$ and $\#_{end}$, and the respective start and end times for each window.
    In the case of finetuning a model (e.g. \chronoBlip or \chronoQwen), we freeze the MLLM and only finetune additional adapter layers leveraging parameter-efficient finetuning \cite{hu2021lora}.
  }
  \label{fig:architecture}
  
\end{figure*}

To explore how to best represent time in an MLLM, we first select the task of moment retrieval as a suitable test bed.
The task is to temporally localize all relevant moments in an untrimmed video given an open-ended natural language query.
Therefore, a key challenge is to effectively model the contextual and temporal relationship between the different events in the video, for example, as illustrated in \cref{fig:architecture}, a man is talking to the camera followed by him watching a show.
Furthermore, due to the amount of information in a video, it is computationally unfeasible to leverage all frames of a video as context in a single forward pass.
To tackle these challenges, we cast moment retrieval as a language modeling task and leverage the contextual understanding ability of generative MLLMs to interpret and comprehend the semantic and temporal action development in frames that represent a video.
We first thoroughly explore the design space of timestamps in MLLMs.
This analysis resulted in the proposed recipe, \chrono, that can enable image-text pretrained MLLMs to interpret video frames and their respective temporal grounding in both a finetuning and even zero-shot setting.
In \cref{sec:Model}, we present the proposed blueprint, \chrono.
In \cref{sec:Training} we discuss how the model-agnostic \chrono blueprint can be leveraged for adapting an MLLM using finetuning and a zero-shot setup.

\subsection{\chrono blueprint}
\label{sec:Model}
This work investigates how to enable image-text pretrained MLLMs to comprehend video and time for temporal grounding tasks in videos given a natural language query.
To do so, we cast the traditional moment retrieval task as an open-ended sequence-to-sequence problem, where we design a novel multimodal input sequence that contains (a) the visual semantic context given the sampled frames of the video, (b) the temporal context for each event, which we model using the temporal location of each frame (i.e., the timestamp) as well as the total video duration, and (c) the query in the form of a natural language description.
The model outputs a time interval sequence where each time interval represents a moment associated with the query and follows the formatting of a nested list of (potentially) multiple moments with a start and end time.
Notably, we do not add any new special tokens and use the vocabulary native to the backbone MLLM, which makes the framework more general and applicable in a zero-shot setting.

To model contextual and temporal relationships between the different actions in a video and a natural language query, we cast moment retrieval as a sequence-to-sequence task and develop a novel multimodal input sequence.
As illustrated in \cref{fig:architecture}, it consists of frames $f_n$, timestamps $t_n$ ($n=1,...,F$), video duration $d$, query $q$, and task prompt $p$ stating the task at hand and, in the zero-shot setting, a format-adherence prompt, which we show in Appendix \cref{fig:Supp:GPT4o_prompt_1}.
Notably, there are multiple ways of representing timestamps $t_n$.
Time can be represented in absolute or relative form, e.g., ``79.9'' seconds or ``0.40'' for a relative position (e.g. 0 being the start of the video and 1 the end); it could be a decimal number or be rounded to the nearest full integer value, e.g., ``80'' seconds or ``40'' for a relative position; the timestamps could be concatenated after the frames $f_n$ or \emph{interleaved} with them.
As illustrated in \cref{fig:architecture}, we find that representing time as simple text tokens of whole seconds (integer absolute form) and interleaving them with corresponding frames yields the best performance.
This finding leads to the final design for the multimodal input sequence
\begin{equation}
    x = [f_1, r(t_1), f_2, r(t_2), ..., f_F, r(t_F), d, q, p]
\end{equation}
where $r(\cdot)$ represents the rounding operation.
In \cref{sec:ablations}, we explore the different variations of representing time and concatenation of the different sequence elements of the input.

Finally, the model is tasked to predict the sequence of potentially multiple relevant windows $m$, which follows the formatting of a nested list of moments with a start and end time in seconds:
\begin{equation}
    y = [[t_{start}^1, t_{end}^1], [t_{start}^2, t_{end}^2], ...]
\end{equation}

Notably, this blueprint is MLLM-architecture-agnostic and can be used to adapt any image-text pretrained model through both simple finetuning and even in a zero-shot setting, which we discuss in \cref{sec:Training}.

\subsection{Training and Zero-Shot \chrono Setup}
\label{sec:Training}
In this section, we describe two different options to apply the \chrono blueprint.
We can either train the generative MLLM or perform zero-shot adaptation on the task of moment retrieval. 
We cast the traditional computer vision problem of moment retrieval as an open-ended natural language sequence-to-sequence task.

\subsubsection{Training Setup}
To train the model, we optimize for the standard maximum likelihood objective.
In order to make finetuning less compute intensive, we employ parameter-efficient finetuning via LoRA~\cite{hu2021lora} and randomly sample video frames.
This allows us to only train a fraction of the parameters of the MLLM. 
We use a rank of 8 and apply it to all linear layers in the large language model, yielding about 19 million trainable parameters for \chronoBlip.
Additionally, by randomly sampling the frames, we optimize those weights by analyzing different frames and timestamps of the same videos each epoch.
In \cref{sec:Supp:trainableparams} and \cref{sec:Supp:number_of_frames}, we study the effect of the number of trainable parameters and the number of frames, respectively.
We provide more details in \cref{sec:Setup_Implementation_Details}.

To demonstrate the applicability of the \chronoWatch{} blueprint to different model architectures, we explore both finetuning BLIP-2~\cite{li2023blip2} and Qwen2.5-VL~\cite{Qwen2.5-VL}. 
BLIP-2~\cite{li2023blip2} is an encoder-decoder style MLLM, whereas Qwen2.5-VL~\cite{Qwen2.5-VL} is a decoder-only model.

\myparagraph{Finetuning BLIP-2}
For BLIP-2, we use its frozen image encoder combined with Q-Former as the general-purpose frame encoder.
The frame encoder is separately applied to each of the $F$ sub-sampled frames, generating the frame embeddings and projecting them into the language space, yielding a shared embedding space.
Prior works~\cite{Wang2024InternVideo2SF, Ren2023TimeChatAT, yang2023vid2seq, zhang2023videollama} incorporate further trainable transformer-based modules to generate temporally and contextually correlated frame- or video-level embeddings.
In contrast, we solely leverage the self- and cross-attention mechanisms of the LLM to learn the temporal and contextual relationships between frames.
We coin this instantiation of the \chrono blueprint \chronoBlip.

\myparagraph{Finetuning Qwen2.5-VL}
We explore finetuning two models of the Qwen2.5-VL~\cite{Qwen2.5-VL} family, the 3 and 7 billion parameter versions.
Qwen2.5-VL jointly finetunes a ViT encoder and an LLM decoder.
The ViT encoder applies full attention across all patches of all images in some layers if a succession of images is embedded.
However, if two images are separated by some text, like in interleaved timestamp setting, the images are encoded by the ViT independently.
We denote applying the \chrono blueprint to Qwen2.5-VL \chronoQwen. 
Notably, we also show that Qwen2.5-VL can be used in a zero-shot manner for moment retrieval, although its performance is significantly lower in comparison to when finetuned.

\subsubsection{Zero-Shot Setup}
For zero-shot adaptation, we can apply the \chrono framework to an MLLM with instruction-following abilities by providing it with the video frames interleaved with their respective timestamps, the video duration, user query, and instruction prompt, just as in the training setup.
We observe that additional format prompting helps the MLLM in this zero-shot setting.

For the zero-shot experiments, we aim to demonstrate the ability of the \chrono blueprint on a model which we do not finetune for moment retrieval.
Most zero-shot experiments use GPT-4o~\cite{GPT4o} since it is a stronger model.

We call this instantiation of the framework \chronoGPT and provide the full prompt used in Appendix \cref{fig:Supp:GPT4o_prompt_1}.
We also demonstrate the advantage of using the \chrono blueprint in a zero-shot setup using Qwen2.5-VL~\cite{Qwen2.5-VL} models in \cref{sec:qwen_abl}.

\section{Experiments}
\label{sec:Experiments}

This section describes the effectiveness of the design choices in this work and compares the proposed method to the state-of-the-art.
We validate \chrono on the three most widely used video moment retrieval (MR) datasets Charades-STA~\cite{gao2017charadesSTA}, QVHighlights~\cite{lei2021qvhighlights}, and ActivityNet Captions~\cite{krishna2017ActivityNetCaptions}, and extend the framework to the task of grounded video question answering on the NExT-GQA~\cite{Xiao2023NExT-QGA} benchmark.

We present and analyze the experiments in this section, as well as introduce the benchmarks used and metrics studied.
We first compare to SOTA methods, then present qualitative results, and finally ablate key design choices.
\subsection{Benchmarks}
\subsubsection{Moment Retrieval}
\textbf{Charades-STA~\cite{gao2017charadesSTA}} includes 9,848 videos with an average duration of 30.6 seconds.
The dataset contains 16,128 annotations with an average moment length of 8.1 seconds and an average query length of 7.22 words.
The dataset is originally divided into two splits: training (12,408) and test (3,720).
To avoid overfitting on the test set during training, we designate a part of the training set as a new validation set.
The videos used in the validation set are not contained in the training set.
The new dataset split consists of a split for training, validation, and testing, with 11,166, 1,242, and 3,720 annotations, respectively.
For a fair comparison, after completing the ablations, we train the final model on the original training set and report numbers on the test set.
We will share this split for reproducibility.

\textbf{QVHighlights~\cite{lei2021qvhighlights}} is one of the most recent benchmarks and contains 10,148 videos with a duration of 150 seconds.
The videos were cropped out of YouTube videos, and each video is annotated with at least one query with an average length of 11.3 words describing the relevant moment.
The target windows have an average length of 24.6 seconds.
The dataset is split into training, validation, and test sets with 7,218, 1,150, and 1,542 queries, respectively.
This benchmark is challenging because one query can be associated with multiple moments in a video.
The test set targets are withheld and thus guarantee a fair benchmark.
The evaluation for the test split can only be measured through submitting the prediction to the evaluation server\footnote{Eval server: \href{https://codalab.lisn.upsaclay.fr/competitions/6937}{https://codalab.lisn.upsaclay.fr/competitions/6937}}.

\textbf{ActivityNet Captions \cite{krishna2017ActivityNetCaptions}}  contains 20,000 videos with an average duration of 2 minutes.
The dataset contains 72,000 segments, each human-annotated with a caption that includes, on average, 13.5 words.
The dataset is divided into three splits, train (37,421), val\_1 (17,505), and val\_2 (17,031).
Following \cite{yan2023unloc}, we use the train split for training, val\_1 for validation, and val\_2 for testing.

\subsubsection{Temporally Grounded Question Answering}
\textbf{NExT-GQA~\cite{Xiao2023NExT-QGA}} extends the NExT-QA~\cite{Xiao2021NExTQA} benchmark by providing temporal grounding for the moments in the video, that are relevant for answering the question for the validation and test sets, making this a weakly-supervised dataset.
The dataset contains the original training split of NExT-QA, which includes 34,132 samples.
The validation set contains 3,358 questions with 3,931, meaning a sample can potentially have multiple relevant moments, but where predicting any single one of those moments is a correct prediction, i.e., one does not have to predict all relevant moments.
The test set has 5,553 questions with 6,600 segments.
The segments have an average duration of 7.3 and 6.7 seconds for the validation and test set, respectively.

\subsection{Metrics}
The most commonly used metrics for MR are Recall@K and mean average precision (mAP) computed under different Intersection over Union (IoU) thresholds.
The Recall@K metric is defined as the percentage of the top-K predicted segments having a larger temporal IoU than the threshold with a ground truth segment.
Following the recent development \cite{lee2023bamdetr, lei2021qvhighlights, moon2023CQ_DETR, moon2023QD-DETR}, we report the more challenging $R1@0.5$ and $R1@0.7$ scores, which correspond to the $Recall@1$ scores at IoU thresholds of $0.5$ and $0.7$, respectively, and the mean IoU (mIoU) score.
As in these prior works, for MR on QVHighlights, we report the mAP score at IoU thresholds of $0.5$ and $0.75$ and the average mAP, respectively.

NExT-GQA~\cite{Xiao2023NExT-QGA} introduces the metrics \textit{(i)} \textit{Intersection over Prediction (IoP)}, which is defined as the ratio between the length of the temporal intersection between the predicted relevant moment and ground truth and the length of the predicted moment itself, and \textit{(ii)} \textit{Accuracy@GQA (A@GQA)} which is the accuracy of correctly answered questions that have a grounding score of $IoP>0.5$.
The IoP loosens the common IoU metric by only considering the intersection over the length of the prediction instead of over the union.
This metric encourages very short-moment predictions and is not concerned with a precise coverage of the entire relevant moment.
Therefore, following prior work, we report mean IoU (mIoU) together with IoP in NExT-GQA.

\subsection{Comparison to the SOTA in Moment Retrieval}
\label{sec:comparisonToSOTA}

\begin{table*}[tb]
  \centering
  \caption{
    \textbf{Comparison to state-of-the-art methods in moment retrieval (left) and grounded video question answering (right).}
    Bold numbers indicate best result for a given benchmark and metric among all methods.
   }
  \label{tab:mr_and_gqa_sota}
\makebox[\textwidth][c]{%
  \begin{subtable}[t]{0.63\textwidth}
    \centering
    \caption{
        \textbf{Moment Retrieval on QVHighlights, Charades-STA, and ActivityNet Captions test sets}.
        For QVHighlights, $^{\dagger}$ indicates validation set was used.
    }
    \label{tab:SOTAcomparison}
    \renewcommand{\tabcolsep}{2.35pt}
\begin{tabular}{l@{\quad}ccccc@{\quad}cc@{\quad}cc@{}}
        \toprule
        & \multicolumn{5}{c}{\textbf{QVHighlights}} & \multicolumn{2}{c}{\textbf{Charades-STA}} & \multicolumn{2}{c}{\textbf{ActivityNet}} \\
        \cmidrule(lr){2-6} \cmidrule(lr){7-8} \cmidrule(l){9-10}
        \textbf{Method} & R1@.5 & R1@.7 & mAP & mAP@.5 & mAP@.75 & R1@.5 & R1@.7 & R1@.5 & R1@.7 \\
        \midrule
        \multicolumn{10}{@{}l}{\textit{\textbf{Finetuned for Moment Retrieval}}} \\
        VLG & -- & -- & -- & -- & -- & -- & -- & 46.30 & 29.80 \\
        SeViLa & 54.50 & 36.50 & -- & -- & -- & -- & -- & -- & -- \\
        CG-DETR & 65.43 & 48.38 & 42.90 & 64.51 & 42.77 & 58.44 & 36.34 & -- & -- \\
        UnLoc-L & -- & -- & -- & -- & -- & 60.80 & 38.40 & 48.30 & 30.20 \\
        EaTR$^{\dagger}$ & 61.36 & 45.79 & 41.74 & 61.86 & 41.91 & 68.47 & 44.92 & -- & -- \\
        InternVideo2-6B & 71.42 & 56.45 & 49.24 & -- & -- & \textbf{70.03} & 48.95 & -- & -- \\
        BLIP-2 & 69.10 & 46.52 & 37.87 & 60.68 & 38.93 & 43.33 & 32.60 & 25.84 & 9.72 \\
        \chronoQwenWatch & {73.48} & {55.10} & 42.83 & {63.74} & {44.47} & {67.98} & {45.22} & {51.25} & {32.79} \\
        \chronoBlipWatch & \textbf{74.77} & \textbf{60.51} & \textbf{51.37} & \textbf{68.12} & \textbf{53.38} & 69.31 & \textbf{49.29} & \textbf{53.92} & \textbf{35.55} \\
        \midrule
        \multicolumn{10}{@{}l}{\textbf{\textit{Zero-Shot}}} \\
        GPT-4o$^{\dagger}$ & 5.49 & 2.17 & 1.10 & 2.81 & 0.76  & 7.02 & 2.12 & 7.20 & 2.80 \\
        \chronoGPTWatch$^{\dagger}$ & \textbf{61.68} & \textbf{41.80} & \textbf{33.01} & \textbf{52.39} & \textbf{35.05}  & \textbf{28.76} & \textbf{10.99} & \textbf{31.06} & \textbf{17.51} \\
        \bottomrule
    \end{tabular}

  \end{subtable}

  \hfill
  \begin{subtable}[t]{0.35\textwidth}
    \centering
    \caption{
        \textbf{Grounded Video Question Answering on NExT-GQA test set.} 
        First four models are known to be explicitly finetuned on NExT-QA.
        See discussion in \cref{sec:next-gqa}.
    }
    \label{tab:NExTGQA}
    \renewcommand{\tabcolsep}{2.35pt}
\begin{tabular}{l r r r r}
    \toprule
    \textbf{Method} & \textbf{mIoU} & \textbf{IoP@.5} & \textbf{A@GQA} & \textbf{A@QA} \\ 
    \midrule
    
    \multicolumn{5}{@{}l}{\textbf{\textit{Finetuned on Dataset}}}\\
    SeViLA~\cite{yu2023SeViLA} & 21.7 &  22.9 & 16.6 & 68.1 \\
    FrozenBiLM~\cite{Xiao2023NExT-QGA} & 9.6 & 23.7 & 17.5 & 70.8 \\
    \chronoBlipWatch & \textbf{28.7}  & \textbf{24.6} & \textbf{19.4} & \textbf{73.9} \\
    \midrule
    \multicolumn{5}{@{}l}{\textbf{\textit{Zero-Shot}}}\\
    Uniform 60F & 21.1 & 21.1 & 8.7 & 7.1 \\
    LLoVi~\cite{Zhang2023LLoVi} & 21.5 & 38.0 & 26.8 & \multicolumn{1}{c}{-} \\
    DeVi~\cite{Qin2024DeVi} & 22.3 & 37.9 & 28.0 & 71.6 \\
    \chronoGPTWatch & \textbf{36.5} & \textbf{50.8} & \textbf{42.1} & \textbf{79.3}  \\
    \bottomrule
\end{tabular}

  \end{subtable}}
\end{table*}

We compare \chrono to the state-of-the-art (SOTA) approaches.
Specifically, we compare \chronoBlip and \chronoQwen to other finetuned baselines and \chronoGPT to zero-shot baselines.

Firstly, in \cref{tab:SOTAcomparison}, we include a comparison to a vanilla BLIP-2 baseline (frames-only, without the \chrono sequence design), where we finetune and evaluate it on each dataset, respectively.
As expected, the \chronoBlip model significantly outperforms this baseline.
Notably, for QVHighlights, this improvement is less pronounced.
We hypothesize this is due to the videos in this benchmark having constant duration.
Nevertheless, it still performs far worse than the \chronoBlip model.

Compared to prior SOTA methods on QVHighlights (QVH), \chronoBlip outperforms the previous SOTA InternVideo2~\cite{Wang2024InternVideo2SF} on all metrics.

On Charades-STA, the \chronoBlip model also achieves a new state-of-the-art, improving over the previous SOTA model~\cite{Wang2024InternVideo2SF} for R1@0.7.
While InternVideo2 achieves good performance by extracting intermediate features and leveraging CG-DETR~\cite{moon2023CQ_DETR} as a localization head, we demonstrate strong results with a fraction of the training data, smaller model size, fewer trained parameters, and while not relying on a task-specific output head.
\chronoBlip also outperforms the previous SOTA \cite{yan2023unloc} in moment retrieval on ActivityNet Captions by 5.62\% and 5.35\% on R1@0.5 and R1@0.7, respectively.

The \chrono model achieves performance that is 20.14\% and 23.29\% higher than SeViLa on QVHighlights in Recall@1 at thresholds of IoU=0.5 and IoU=0.7, while using the same backbone, BLIP-2~\cite{li2023blip2}.
This demonstrates the effectiveness of the novel multimodal sequence design.
In contrast, SeViLa classifies each frame as relevant to the query or not without any contextual and temporal information about the video.

The results for \chronoQwen follow similar trends to those of \chronoBlip, indicating that the \chrono blueprint generalizes beyond a single backbone to decoder-only architectures of different sizes and training recipes.

Moreover, we also evaluate the \chronoGPT model on all three MR benchmarks in a zero-shot setting.
As expected, the zero-shot performance achieved by \chronoGPT is lower compared to methods explicitly trained for Moment Retrieval. 
However, \chronoGPT performs significantly better than the baseline GPT-4o. 
We observe that without any training or timestamps, GPT-4o completely fails to associate specific frames with a specific time, even when provided with the total duration and the information that the frames are sampled uniformly.
We suspect the temporal encoding and training task distribution significantly affect this ability.
Therefore, in \cref{tab:qwen_abl}, we show results with Qwen-2.5-VL models in the zero-shot setting, which employ a positional encoding method that explicitly represents the time associated with each frame.

Prior work~\cite{qian2024momentor, Ren2023TimeChatAT, huang2024vtimellm} performs extensive video-language instruction finetuning on multiple temporal grounding datasets and then evaluates on a held-out target dataset (commonly Charades-STA).
To compare under a clear protocol, we report cross-dataset transfer results in \cref{tab:transfer_results}, which we define as finetuning on a source dataset and evaluating on a different target dataset without any further tuning on the target.
Concretely, we use two pairs: ActivityNet Captions $\rightarrow$ Charades-STA, and QVHighlights $\rightarrow$ ActivityNet Captions, keeping the same training setup and frame counts as in the main experiments. 
Despite training on a single source dataset (versus broad instruction-tuning in prior work), \chronoBlip achieves competitive or superior transfer performance.

This demonstrates the surprising effectiveness of simple but deliberate design choices for leveraging an image-text pretrained MLLM to achieve state-of-the-art results.

\FloatBarrier
\begin{table}[tb]
    \renewcommand{\tabcolsep}{5pt}
    \centering
    \caption{\textbf{Transfer Learning Results} for moment retrieval across datasets. $^*$ indicates that Momentor was trained on a subset of ActivityNet Captions.}
    \label{tab:transfer_results}
    \begin{tabular}{l@{\quad}cc@{\quad}cc@{}}
        \toprule
        & \multicolumn{2}{c}{\textbf{Charades-STA}} & \multicolumn{2}{c}{\textbf{ActivityNet}} \\
        \cmidrule(lr){2-3} \cmidrule(l){4-5}
        \textbf{Method} & R1@.5 & R1@.7 & R1@.5 & R1@.7 \\
        \midrule
        Momentor$^*$~\cite{qian2024momentor} & 26.60 & 11.60 & 23.00 & 12.40 \\
        TimeChat~\cite{Ren2023TimeChatAT} & 32.20 & 13.40 & -- & -- \\
        VTimeLLM~\cite{huang2024vtimellm} & 34.30 & 14.70 & -- & -- \\
        \chronoBlipWatch & \textbf{34.87} & \textbf{18.04} & \textbf{28.64} & \textbf{16.44} \\
        \bottomrule
    \end{tabular}
\end{table}

\subsection{Grounded Video QA on NExT-GQA}
\label{sec:next-gqa}

Next, we challenge \chrono to a new task, grounded video question answering (GQA) on the NExT-GQA~\cite{Xiao2023NExT-QGA} benchmark, i.e., localizing a moment that is relevant to answering a question and, then even answering the question.

\subsubsection{Localize-then-answer using \chronoBlip}
To evaluate \chronoBlip on GQA, we follow the localizer-answerer approach of SeViLa~\cite{yu2023SeViLA} using 60 frames for localizing the relevant moment, and then resampling 60 new frames out of that segment for answering the question.
We apply the QVH-pretrained \chronoBlip model as the localizer and finetune a separate Flan-T5 XL LLM as the answerer.
Notably, we \textbf{do not finetune} the localization model, thus applying it to a new domain, the domain of questions, where a relevant moment does not have to be explicitly described in the question to be relevant to answering it. 
We also compare to uniformly sampling 60 frames. 
\cref{tab:NExTGQA} demonstrates the performance of \chronoBlip on the NExT-GQA~\cite{Xiao2023NExT-QGA} benchmark and compares it to prior state-of-the-art models, SeViLa and FrozenBiLM (NG+)~\cite{Xiao2023NExT-QGA}.

Inspecting the mIoU scores, we can see that we outperform the FrozenBiLM model by $19.13\%$.
In fact, on the mIoU, FrozenBiLM and SeViLa are performing worse or on par with the uniform baseline, respectively.
Both \chronoBlip (localizer) and SeViLa's localizer were pretrained solely on QVH.
Yet, \chronoBlip adapts much better to the new question-based setting by achieving a $7.03\%$ higher mIoU than SeViLa.
Finally, \chronoBlip outperforms the prior SOTA by $1.90\%$ on the A@GQA metric. 

\subsubsection{Single localizer-answerer using \chronoGPT}

Following the same protocol, we evaluate the \chronoGPT model in a zero-shot setting.
The timestamp design enables strong zero-shot moment retrieval for \chronoGPT (see \cref{sec:ablations} for the ablations).
Paired with the strong QA abilities of a foundational model, we achieve a new state-of-the-art result on NExT-GQA over prior work~\cite{Zhang2023LLoVi, Qin2024DeVi}, which also leverages GPT-4 and GPT-4o, respectively.

The best \chronoGPT setup for NExT-GQA uses a single stage, i.e., returns answer and temporal grounding in the same model response. 
We find that this performs better than 2 stages, where the model first performs moment retrieval, and then the model is asked to answer the question. 
This is the case even when the second stage is able to zoom in on the window, use more frames overall, and keeps the first stage in context. 
We choose to use 1.4FPS for the final results. This offers 1-5\% relative improvement in all metrics compared to using 60 frames for every video.
This uses an equivalent number of frames across the entire dataset (with an average length of around 40 seconds) compared to using 60 frames for every video. 
Using absolute interleaved timestamps significantly improves temporal grounding and, therefore, grounded accuracy. 
We find that grounding modestly helps overall accuracy, i.e., not thresholded by grounding, as can be seen also in \cref{tab:supp:abl_GPT4o_timestamps_nextQA}. 
This could indicate that including temporal grounding capabilities into models that process video cannot only offer greater interpretability but also improve their overall performance at other tasks like video QA.
Additionally, this indicates that the increase in grounded accuracy is not strongly driven by better answering capabilities, but truly better temporal localization.
\begin{table}[tb]
    \renewcommand{\tabcolsep}{4pt}
    \caption{\textbf{Does moment retrieval help  Video QA?} 
    Asking the model to ground its answer provides modest improvements in accuracy, with grounding accuracy significantly increasing when using \chronoWatch{}.
    For cost reasons, we only perform this ablation on 1000 samples from NExT-GQA validation split, using GPT-4o.
    }
    \label{tab:supp:abl_GPT4o_timestamps_nextQA}
    \centering
    \begin{tabular}{@{}ccccccc@{}}
        \toprule
        \textbf{MR} & \textbf{\chronoWatch{}} & \textbf{mIoU} & \textbf{mIoP} & \textbf{IoP@.5} & \textbf{A@GQA} & \textbf{A@QA} \\
        \midrule
        \xmark     & \xmark      & -  & -  & -  & -  & 79.80 \\
        \cmark & \xmark      & 22.47 & 29.85 & 27.13 & 22.03 & 80.07 \\
        \cmark & \cmark & \textbf{37.02} & \textbf{51.62} & \textbf{52.45} & \textbf{43.35} & \textbf{80.15} \\
        \bottomrule
    \end{tabular}
\end{table}

\subsection{Qualitative Results}
\label{sec:qualitativeResults}

\begin{figure*}[htbp]
  \centering
  \includegraphics[width=\textwidth]{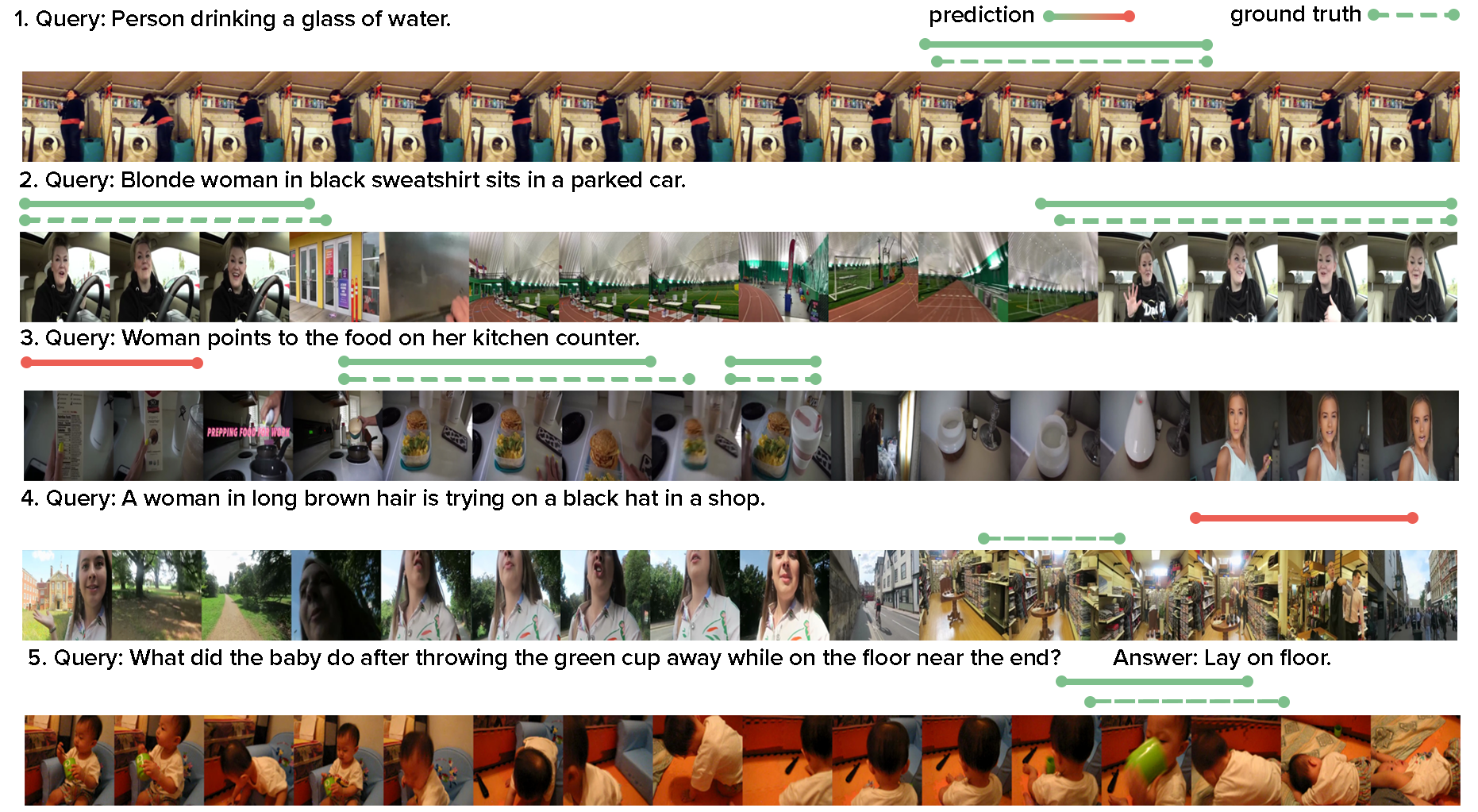}
  \vspace{-10pt}
  \caption{
    (1,2) correct multi-window moment retrieval.
    (3) \chronoBlip predicts two correct moments alongside one that is false.
    (4) \chronoBlip predicts a moment that only partly matches the query. 
    (5) \chronoBlip successfully predicts an out-of-distribution example on NExT-GQA while being trained on QVH.
    (1) is from Charades-STA,  (2) to (4) are from QVHighlights, and (5) is from NExT-GQA.
  }
  \label{fig:qualitative}
  
\end{figure*}

\cref{fig:qualitative} shows qualitative results that demonstrate \chronoBlip's ability to predict multiple windows (2 and 3), generalize to unseen data distributions (5), as well as some failure modes like predicting wrong segments (3) and only partially adhering to the prompt due to low image resolution (4). 
In (4), the query is "A woman in long brown hair is trying on a black hat in a shop". 
\chronoBlip predicts a moment where the woman is clearly visible in a shop but not trying on a hat.
In the ground truth segment, the woman is trying on a hat but is not clearly visible since she is far away.
We provide more videos with localizations in \cref{sec:Supp:Additional_Qualitative_Results}. 

\subsection{Ablation Studies}
\label{sec:ablations}

By default, the \chrono models for Charades-STA (QVHighlights / ActivityNet Captions) leverage 20 (60) visual frames, their associated timestamps, video duration, a query, and a task prompt.
If not stated otherwise, we represent time as simple text tokens of whole seconds and interleave these with frame tokens.
In the following, we ablate the impact of these design choices on the downstream moment retrieval performance by reporting results for \chronoBlip on the Charades-STA validation set and for \chronoGPT on the QVHighlights validation set.
Moreover, we provide further evidence for the generalization of the blueprint by performing ablations on the Qwen2.5-VL~\cite{Qwen2.5-VL} family of models across different model sizes in zero-shot and finetuned settings.
In several of the ablation experiments, we also report mIoU for completeness, although we find the relative performance in terms of mIoU follow those of R1@0.5, which is, in any case, more informative of demanding temporal grounding criteria. 
Therefore, in many other cases we report only R1@0.5 and R1@0.7.

\begin{table}[tb]
\setlength{\tabcolsep}{4pt}
   \caption{
      \textbf{Ablation studies on input sequence and timestamp design.}
      For finetuned \chronoBlip, we ablate on Charades-STA, and for zero-shot \chronoGPT, on QVHighlights.
      We use validation set in all cases. 
      D: Video Duration, T: Frame Timestamps, Rep: Representation, Rel: Relative, Abs: Absolute, Prec: Precision, Dec: Decimal, Int: Integer, Inter: Interleaved.
      Bold indicates best result for a given base model and metric.
      Underlined indicates second best.
   }
   \label{tab:ablations}
   \centering
   \begin{tabular}{@{}cccc@{\hspace{8pt}}|@{\hspace{8pt}}cc@{}} 
       \multicolumn{6}{c}{\textbf{(a) Prompt design components}} \\
       \toprule
       & & \multicolumn{2}{c}{\textbf{\chronoBlip}} & \multicolumn{2}{c}{\textbf{\chronoGPT}} \\
       \cmidrule(lr){3-4} \cmidrule(lr){5-6} 
       \textbf{D} & \textbf{T} & R1@.5 & R1@.7 & R1@.5 & R1@.7 \\
       \midrule
       \xmark & \xmark & 43.84 & 23.95 & 5.74 & 1.94  \\
       \cmark & \xmark & 55.60 & 32.63 & 4.39 & 1.87   \\
       \xmark & \cmark & \textbf{67.81} & \underline{43.33} & \underline{60.00} & \underline{41.68} \\
       \cmark & \cmark & \underline{67.28} & \textbf{46.70} & \textbf{61.68} & \textbf{41.80} \\
       \bottomrule
   \end{tabular}\\[8pt] 
   \begin{tabular}{llcccc@{\hspace{8pt}}|@{\hspace{8pt}}cc}
       \multicolumn{8}{c}{\textbf{(b) Timestamp design}} \\
       \toprule
       & & & & \multicolumn{2}{c}{\textbf{\chronoBlip}} & \multicolumn{2}{c}{\textbf{\chronoGPT}} \\
       \cmidrule(lr){5-6} \cmidrule(lr){7-8} 
       \textbf{\#} & \textbf{Rep.} & \textbf{Prec.} & \textbf{Inter.} & R1@.5 & R1@.7 & R1@.5 & R1@.7 \\
       \midrule
       (1) & Rel & Dec & \xmark & 62.30 & 37.22 & 48.39 & 26.06 \\
       (2) & Abs & Dec & \xmark & 62.38 & 36.33 & 44.19 & 24.58 \\
       (3) & Rel & Int & \xmark  & 63.10 & 36.01 & 30.42 & 19.90 \\
       (4) & Abs & Int & \xmark & 64.39 & 41.80 & 36.77 & 20.68 \\
       (5) & Rel & Int & \cmark & \underline{65.19} & \underline{44.94} & \textbf{62.84} & \textbf{42.19} \\
       (6) & Abs & Int & \cmark & \textbf{67.28} & \textbf{46.70} & \underline{61.68} & \underline{41.80} \\
       \bottomrule
   \end{tabular}
\end{table}

\subsubsection{Input sequence design}
\label{sec:PromptDesign}
In \cref{tab:ablations} (a), we analyze the effectiveness of each part of the multimodal input sequence of the MLLM by cumulatively adding each component, starting with only providing the frames.
The query is always part of the sequence.
For \chronoBlip, adding the video duration, in addition to the visual inputs, improves the downstream performance significantly (row 2 vs. row 1).
This shows the importance of providing the model a point of reference to infer the temporal location of each frame and demonstrates the ability of the MLLM to learn this association.
ChronoGPT, on the other hand, does not benefit as strongly from the additional information as the fine-tuned BLIP-2 model.
Again, note that with a fixed number of uniformly sampled frames and the video duration, the model has all the information necessary to compute at which timestamp each frame was sampled.
This experiment shows that GPT-4o appears not to be able to leverage this information.

Yet, providing the specific timestamps at which each frame was sampled in an interleaved manner (as illustrated in \cref{fig:architecture}) improves the model performance of both models by a significant margin (rows (3, 4) vs. rows (1,2)).
This demonstrates the relevance of designing the input context to provide as much relevant information as possible while still not relying on additional computation, such as generating and leveraging transcripts.
Providing all timestamps along with the video duration yields the best performance (row 4 vs. 3) for both models.

\begin{table}[tb]
  \caption{\textbf{Input components for zero-shot moment retrieval using GPT-4o} on Charades-STA and QVHighlights val sets.}
  \label{tab:AblationPromptDesign-GPT4o}
  \centering
  \begin{tabular}{@{}l@{\hskip 0.25pt} 
                  S[table-format=2.2]@{\hskip 0.25pt} 
                  S[table-format=2.2]@{\hskip 0.25pt} 
                  S[table-format=2.2]@{\hskip 0.25pt}
                  S[table-format=2.2]@{\hskip 0.25pt} 
                  S[table-format=2.2]@{\hskip 0.25pt} 
                  S[table-format=2.2]@{}}
    \toprule
                     & \multicolumn{3}{c}{\textbf{Charades-STA}} & \multicolumn{3}{c}{\textbf{QVHighlights}} \\
    \cmidrule(lr){2-4} \cmidrule(lr){5-7}
    \textbf{Input}              &
    \multicolumn{1}{c}{R1@.5} &
    \multicolumn{1}{c}{R1@.7} &
    \multicolumn{1}{c}{mIoU} &
    \multicolumn{1}{c}{R1@.5} &
    \multicolumn{1}{c}{R1@.7} &
    \multicolumn{1}{c}{mIoU} \\
    \midrule
    Frames only      & 8.62 & 2.67 & 14.74 & 5.49 & 2.17 & 10.56 \\
    + Duration       & 11.56 & 4.59 & 18.25 & 4.32 & 1.87 & 8.96 \\
    + Timestamps & \textbf{28.76} & \textbf{10.99} & \textbf{35.43} & \textbf{61.68} & \textbf{41.80} & \textbf{57.12} \\
    \bottomrule
  \end{tabular}
\end{table}

Complementary to \cref{tab:ablations} (a), \cref{tab:AblationPromptDesign-GPT4o} isolates the zero-shot GPT-4o setting on Charades-STA and QVHighlights.
Starting from frames only, adding the video duration yields only minor gains, whereas including interleaved timestamps in \chronoGPTWatch{} leads to large improvements across R1@0.5/0.7 and mIoU.
This mirrors the behavior observed for the finetuned \chronoBlip model, reinforcing that the interleaved timestamp design is crucial even for powerful zero-shot MLLMs such as GPT-4o.

\subsubsection{Design of timestamps}
\label{sec:timestamp_design}
\textbf{Numerical precision of timestamps.}
A central question for this work is how MLLMs can best represent and reason about timestamps, as they have to be interpreted in the input and predicted in the output.
We compare different options for representing the timestamps and their impact on the final downstream model performance.
We explore using relative positions w.r.t. the video duration vs. representing each timestamp as absolute time. We compare representing these as decimal numbers (absolute ``79.9'' seconds, relative ``0.40'') vs. integers (absolute ``80'' seconds, relative ``40''). 
In these experiments, the format of the timestamps in model input and output is always consistent. 
In the case of relative positions, we post-process the output to yield an absolute value to compare to the ground truth for the final evaluation.

Inspecting the \chronoBlip ablations in \cref{tab:ablations} (b), we observe that representing the timestamps in decimal form results in lower performance for both absolute and relative form, see rows (1,2) vs. rows (3, 4).
We hypothesize this is the case because of how decimal numbers are tokenized.
Although using decimal timestamps allows for higher temporal resolution, we hypothesize that this lower performance can be attributed to limitations related to the tokenization of decimal numbers.
Depending on the tokenizer, a decimal number such as ``79.9'' or ``42.05'' can be divided into a different number of tokens each, whereas integers (up to a certain number) are tokenized as a single token.
Nevertheless, it is important to note that this behavior varies across tokenizers and can, therefore, be subject to variance.
Further, we observe that absolute position in seconds yields better performance than the relative representation when using integer precision. 
This shows the importance of designing the input and outputs with the representation of the MLLM in mind.

When analyzing the trends for \chronoGPT, in \cref{tab:ablations} (b) and, more detailed, in \cref{tab:supp:abl_GPT4o_timestamps_combined}, we find that these trends are not as evident.
We observe that there is higher variance when evaluating \chronoGPT on the downstream tasks and that it seems, when inspecting only the first four rows, that GPT-4o prefers decimal precision.
We hypothesize that GPT-4o, being a more advanced model, might be more suitable for processing numbers with decimal precision.

\begin{table}[tb]
    \centering
    \caption{\textbf{Ablation: timestamp design for GPT-4o zero-shot on Charades-STA and QVHighlights validation sets.} We compare relative vs. absolute representations, decimal vs. integer precision, and interleaving vs. appending timestamps.}
    \label{tab:supp:abl_GPT4o_timestamps_combined}
    \setlength{\tabcolsep}{2.25pt}
    \begin{tabular}{lcccccccccc}
    \toprule
      &  &  &  &  & \multicolumn{3}{c}{\textbf{Charades-STA}} & \multicolumn{3}{c}{\textbf{QVHighlights}} \\
    \cmidrule(lr){6-8} \cmidrule(lr){9-11}
    \textbf{\#} & \textbf{Rep.} & \textbf{Prec.} & \textbf{Inter.} &  & R1@.5 & R1@.7 & mIoU & R1@.5 & R1@.7 & mIoU \\
    \midrule
    (1) & Rel & Dec & \xmark & & 32.85 & 11.38 & 37.78 & 48.39 & 26.06 & 46.45 \\
    (2) & Abs & Dec & \xmark & & 32.38 & 13.30 & 36.99 & 44.19 & 24.58 & 43.82 \\
    (3) & Rel & Int & \xmark & & 34.68 & \textbf{15.46} & 38.04 & 30.42 & 14.90 & 33.91 \\
    (4) & Abs & Int & \xmark & & 29.91 & 11.28 & 33.43 & 36.77 & 20.68 & 37.41 \\
    (5) & Rel & Int & \cmark & & \textbf{34.86} & 11.20 & \textbf{38.44} & \textbf{62.84} & \textbf{42.19} & \textbf{57.91} \\
    (6) & Abs & Int & \cmark & & 28.35 & 11.29 & 35.43 & 61.68 & 41.80 & 57.12 \\
    (7) & Abs & Dec & \cmark & & 28.72 & 10.74 & 34.78 & 58.77 & 40.16 & 55.83 \\
    \bottomrule
    \end{tabular}
\end{table}

\cref{tab:supp:abl_GPT4o_timestamps_combined} focuses on ablating the timestamp design when using GPT-4o zero-shot for moment retrieval. 
The QVHighlights results support the main hypothesis that interleaving timestamps between frames leads to better localization, also in the zero-shot case. 
The effect of interleaving is less clear with Charades-STA. 
We hypothesize that this difference is caused by the fact that we use only 20 frames (out of shorter videos) for Charades-STA, whereas in the QVHighlights, GPT-4o observes 60 frames (of videos up to several minutes long).
This means appended timestamps are further away from the frame that they refer to, making the gap larger relative to interleaving on QVHighlights.
However, in Charades-STA, the shorter distance between token frames and appended timestamps seemingly has a smaller effect.
Additionally, we could hypothesize that successive frames are more in distribution for GPT-4o than a series of frames with interleaved timestamps, therefore offering some benefit to simply appending the timestamps in very short frame sequences.
This illustrates how some timestamp decisions can be more beneficial if the model can be finetuned, as shown for \chronoBlipWatch{}.
We can also observe a difference in this zero-shot setting, which uses a stronger model.
Interleaving relative integer timestamps yields better performance than using the absolute temporal representation in seconds.
There are to possible explanations for this:
Firstly, a relative timestamp might be more intuitive without an inherent understanding of absolute time (seconds). 
Secondly, the model is able to output predictions in a higher temporal resolution. 
Lastly, the best configurations for both datasets use  integer precision rather than decimal, even if the latter offers a higher temporal resolution, which is in agreement with the results when finetuning, as shown in \cref{tab:ablations}. 

These findings demonstrate the relevance of deliberate sequence design for different MLLMs in zero-shot and finetuned settings and its effect on enabling MLLMs for temporal localization in videos.

\textbf{Interleaving frame and time tokens.}
Moreover, we explore the design choice of ordering frame and time tokens in \cref{tab:ablations} (b).
In rows (1) through (4), we concatenate all frames together, followed by a concatenation of the timestamps, which are each separated by a separator token (``$>$'').
In rows (5) and (6), we observe the benefit of \emph{interleaving} frame tokens with time tokens when applied to both relative and absolute timestamps.
For both models, \chronoBlip and \chronoGPT, the most evident gain of this sequence design is that interleaving frames and timestamps significantly outperforms the non-interleaved settings.
The experiments demonstrate that representing time in seconds, rounded to the respective nearest integer, in an interleaved sequence design yields the best performance when applied to the BLIP-2 backbone; we use this configuration in the SOTA comparisons (\cref{tab:SOTAcomparison}).
For the \chronoGPT setting, on the other hand, relative and absolute representations are on par.

For simplicity and consistency in subsequent experiments, for \chronoGPT we use the row (6) configuration: absolute integer seconds with interleaved timestamps. We adopt these best settings in the SOTA comparisons (\cref{tab:SOTAcomparison}).

\subsubsection{Extending to different model sizes and model families}

\label{sec:qwen_abl}
\begin{table}[tb]

    \renewcommand{\tabcolsep}{4pt}

    \centering
    \caption{
        \textbf{Qwen2.5-VL-Instruct performance comparison with and without Chrono blueprint.} 
        We evaluate different model configurations on Charades-STA test set and measure recall at IoU thresholds of 0.5 and 0.7. 
        For both model sizes and finetuned/zero-shot settings, \colorbox{lightgray}{the Chrono blueprint (in a grey background)} consistently improves moment retrieval. 
    }
    \label{tab:qwen_abl}
    \begin{tabular}{@{}lcccccc@{}}
        \toprule
        \textbf{Size} & \textbf{Finetuned} & \textbf{Chrono} & \textbf{R1@.5} & \textbf{R1@.7} \\
        \midrule
        \multirow{4}{*}{3B} & \multirow{2}{*}{\xmark} & \xmark & 7.74 & 2.88 \\
         & & \rowbg\cmark & \rowbg\textbf{13.87} & \rowbg\textbf{5.00} \\
        \cmidrule{2-5}
         & \multirow{2}{*}{\cmark} & \xmark & 44.41 & 23.47 \\
         & & \rowbg\cmark & \rowbg\textbf{62.10} & \rowbg\textbf{39.87} \\
        \midrule
        \midrule
        \multirow{4}{*}{7B} & \multirow{2}{*}{\xmark} & \xmark & 22.07 & 9.73 \\
         & & \rowbg\cmark & \rowbg\textbf{31.61} & \rowbg\textbf{14.46} \\
         \cmidrule{2-5}
         & \multirow{2}{*}{\cmark} & \xmark & 48.20 & 27.98 \\
         & & \rowbg\cmark & \rowbg\textbf{67.98} & \rowbg\textbf{45.22} \\
        \bottomrule
    \end{tabular}
\end{table}

To explore different model sizes and the ability of the Chrono blueprint to generalize to even further model architectures, we perform additional ablations using the Qwen2.5-VL~\cite{Qwen2.5-VL} family of models and report numbers on the Charades-STA validation set.

We demonstrate that the \chronoWatch blueprint is useful across Qwen2.5-VL sizes and tuning regimes, as shown in \cref{tab:qwen_abl}.
In all scenarios, finetuned/zero-shot or 3B/7B, the \chronoWatch blueprint consistently improves moment retrieval.

Importantly, these experiments allow us to gauge whether M-RoPE, which encode temporal information per frame, are enough for moment retrieval.
These results prove that \chronoWatch's explicit timestamp representation provides benefits beyond what M-RoPE can achieve independently, although the gap in the zero-shot case is remarkably smaller than for GPT-4o, as seen in \cref{tab:SOTAcomparison}, which is presumably not using such positional encoding.

We find that the Qwen2.5-VL models are more sensitive to changes in the instruction/format prompt, and cannot exploit the use of chain-of-thought. 
This can be explained both by their smaller size, and the nature of their training, i.e. instruction following and not reasoning. 
Therefore, we use the simple training prompt for the Qwen2.5-VL models also in the zero-shot setting.

\subsubsection{Effect of number of trainable parameters}
\label{sec:Supp:trainableparams}
\begin{table}[tb]
  \caption{
    \textbf{Ablation on the finetuning setup for \chronoBlip on Charades-STA validation set.} 
    Using 20 input frames and LoRA rank 8 achieves the best recall at IoU$\geq$0.7. 
  }
  \label{tab:supp:frames_lora}
  \centering
  \makebox[\columnwidth][c]{%
    \begin{subtable}[t]{0.48\columnwidth}
    \renewcommand{\tabcolsep}{4pt}
    \centering
    \caption{
        \textbf{Frame count ablation.}
    }
    \label{tab:supp:number_of_frames}
    \begin{tabular}{@{}ccc@{}}
        \toprule
        \textbf{Frames} & \textbf{R1@.5} & \textbf{R1@.7} \\
        \midrule
        5 & 54.50 & 32.72 \\
        10 & 64.63 & 41.80 \\
        15 & 67.79 & 44.07 \\
        20 & 67.28 & \textbf{46.70} \\
        25 & \textbf{67.95} & 43.35 \\
        30 & 64.58 & 41.51 \\
        \bottomrule
    \end{tabular}
\end{subtable}

    \hfill
    \begin{subtable}[t]{0.48\columnwidth}
  \centering
  \caption{
    \textbf{LoRA rank ablation.}
  }
  \label{tab:supp:lora}
  \begin{tabular}{@{}cccc@{}}
    \toprule
    \textbf{Rank} & \textbf{R1@.5} & \textbf{R1@.7} & \textbf{mIoU} \\
    \midrule
    2 & 66.03 & 43.03 & 56.59 \\
    4 & 65.71 & 43.59 & 56.78 \\
    8 & 67.28 & \textbf{46.70} & \textbf{57.46} \\
    16 & \textbf{67.72} & 45.38 & 57.35 \\
    \bottomrule
  \end{tabular}
\end{subtable}
}
\end{table}

In \cref{tab:supp:lora}, we show the effect of different LoRA ranks and their resulting number of trainable parameters for \chronoBlip.
Note that this is different from scaling the model altogether, as shown in \cref{tab:qwen_abl}, since we keep the total model size constant but vary the model capacity devoted to moment retrieval.
For simplicity, we report results on the Charades-STA validation set with \chronoBlip.
We observe that setting the rank to 16, i.e. training double the number of parameters does not yield any significant improvement (row 4).
Moreover, training fewer parameters, i.e. 10 or 6 million, by setting the rank to 4 or 2, respectively, does degrade the performance of \chrono (rows 1 and 2).

\subsubsection{Effect of number of frames}
\label{sec:Supp:number_of_frames}
In \cref{tab:supp:number_of_frames}, we show the impact of the number of input frames and demonstrate the ability of the image-text pretrained MLLM to adapt to the video modality. 

For Charades-STA with an average video duration of 30 seconds, using 20 frames achieves the highest R1@0.7, and offers a good efficiency-performance trade-off when considering R1@0.5 and mIoU scores.

For QVHighlights, which consists of 150-second long videos, and for ActivityNet, which on average has 120-second long videos, we have found 60 frames to perform the best, showing the ability of \chronoBlip to comprehend a relatively long sequence of interleaved visual and textual tokens.

\section{Conclusion}
\label{sec:Conclusion}
We introduce \chronoWatch, a simple model-agnostic blueprint for representing time in multimodal large language models, thus enabling them to perform tasks such as moment retrieval and grounded video question answering.
We show that originally image-text pretrained \chrono models can easily adapt to the video-language modality and greatly benefit from an input design that incorporates visual and temporal information as a simple interleaved token sequence of visual and text tokens.
Specifically, finetuned \chronoBlip achieves state-of-the-art results on the popular moment retrieval benchmarks, whereas zero-shot \chronoGPT achieves a new SOTA in grounded video question answering.
Through extensive ablations, we find that this simple model-agnostic design outperforms all prior SOTA methods, which rely on complex modifications like task-specific model architectures, extensive video pretraining, or additional input signals, such as video transcripts or novel time embedding modules.
Furthermore, we show that the \chronoWatch blueprint generalizes across training setups (finetuned vs zero-shot), model architectures and sizes (\chronoBlip, \chronoQwen, \chronoGPT) and tasks (moment retrieval, grounded video QA).
We hope this versatile sequence-to-sequence design of \chrono serves as a fundamental baseline or potential best practice for encoding time in video-language models and sparks further research in leveraging image-text pretrained MLLMs for video understanding tasks.

\section{Acknowledgements}
The research was partially funded by a LOEWEStart-Professur (LOEWE/4b//519/05.01.002-(0006)/94), LOEWE-Spitzen-Professur (LOEWE/ 4a//519/05.00.002-(0010)/93), and an Alexander von Humboldt Professorship in Multimodal Reliable AI, sponsored by Germany’s Federal Ministry for Education and Research

\IEEEtriggercmd{\enlargethispage{-10\baselineskip}}
\bibliographystyle{IEEEtranN}
\bibliography{main}

\textbf{Hector G. Rodriguez} received the M.Sc. degree in machine learning and the B.Sc. degree in theoretical physics from University College London, London, U.K.
He is currently pursuing the Ph.D. degree in computer science at the Technical University of Darmstadt, Darmstadt,Germany.
His research interests include multimodal reliable AI and efficient multimodal AI.

\textbf{Boris Meinardus} received the B.Sc. degree in computer engineering in 2021 and the M.Sc. degree in computer science in 2024, both from the Technical University of Berlin, Berlin, Germany.
He is currently a Research Scientist at Sakana AI in Tokyo, Japan. 
His research interests are in open-ended learning with large foundation models.

\textbf{Anil Batra} received the B.Tech. degree in electronics and communication from Punjab Technical University, India, in 2007, the M.Sc. degree in computer science from the International Institute of Information Technology, Hyderabad, India, in 2019 and passed the Ph.D. degree in computer science at the University of Edinburgh, Edinburgh, U.K.
He is currently a Research Fellow at University of Surrey and working on Sign Language videos. 
His research interests include multimodal AI, video understanding, and diffusion models.

\textbf{Anna Rohrbach} received the M.Sc. degree in Applied Mathematics from Odesa I. I. Mechnikov National University (Ukraine) in 2010 and the Ph.D. degree in Computer Science from the Max Planck Institute for Informatics and Saarland University (Germany) in 2017. 
She is a Professor at the Technical University of Darmstadt (Germany), where she leads the Multimodal Grounded Learning group and is part of the Multimodal AI Lab. 
Previously, she was a Research Scientist at the University of California, Berkeley (USA). 
Her research is at the intersection of Computer Vision and Natural Language Processing, including image and video description, visual grounding, visual question answering, text-to-image synthesis, and multimodal fact-checking. 
Prof. Rohrbach received the German Pattern Recognition Award of the German Association for Pattern Recognition (DAGM) in 2023. 
She is a recipient of the prestigious €2M LOEWE Start Professorship from the State of Hesse (2024). 
Anna Rohrbach has served as a Reviewer and Area Chair at many top-tier machine learning conferences (CVPR, ECCV, ICCV, NeurIPS, ACL, EMNLP, ICLR) and serves as a Program Chair of ECCV 2026.

\textbf{Marcus Rohrbach} received the B.Sc. and M.Sc. degrees in computer science from the Technical University of Darmstadt, in 2006 and 2009, respectively, and the Ph.D. degree in computer science from the Max Planck Institute for Informatics and Saarland University, in 2014.
He is a Professor at the Technical University of Darmstadt, where he leads the Multimodal Reliable AI Lab. 
Previously, he was a postdoctoral researcher at the University of California, Berkeley, and a Research Scientist at Facebook AI Research in Menlo Park, CA, USA. 
His research interests include computer vision, computational linguistics, and machine learning, with an emphasis on multimodal learning and reliable AI systems. 
Prof. Rohrbach holds an Alexander von Humboldt Professorship in Multimodal Reliable AI and a LOEWE Spitzen Professorship awarded by the state of Hesse, Germany.

\appendices
\clearpage
\setcounter{page}{1}

\section{Implementation Details}

\label{sec:Setup_Implementation_Details}
In this section, we present the specific finetuning and evaluation details for \chronoBlip and \chronoQwen, as well as further specific zero-shot evaluation elements for \chronoGPT

\subsection{Finetuning Details for BLIP-2 and Qwen2.5-VL}
Considering the relatively small scale of the datasets, we choose not to finetune the entire model but rather leverage parameter-efficient finetuning~\cite{hu2021lora}.
For the BLIP-2 LLM backbone, this amounts to training only 19 million parameters, instead of 3 billion as described in \cref{sec:Training}.

Since the objective is a generative language modeling task, the model can output minor formatting inconsistencies that would invalidate the prediction if not treated otherwise.
Similar to~\cite{yang2023vid2seq}, we post-process the model's outputs by applying heuristics to clean up minor flaws in the prediction. 

For Charades-STA, we extract 20 video frames for each video, which is significantly less than prior work~\cite{yan2023unloc}, which requires 128 frames.
We train for up to 20 epochs with a batch size of 32 samples, using A100-80GB GPUs for training.
For BLIP-2, this takes about 20 hours when using data parallelism split across 4 A100-80GB GPUs.
For QVHighlights and ActivityNet Captions, we extract 60 video frames for each video, which is less than required by prior work~\cite{moon2023CQ_DETR, moon2023QD-DETR}.
We train for up to 50 epochs with an effective batch size of 32 samples.
For BLIP-2, finetuning in a single 8 GPU node with gradient accumulation takes about 170 GPU hours.

We start with a learning rate (lr) of 1e-8, perform linear lr warmup to 3e-4 for 10\% of the total number of iterations, and then apply a cosine lr decay.
We utilize the AdamW~\cite{loshchilov2019adamw} optimizer and sample the frames randomly during training. 

\cref{tab:supp:Hyperparameters} shows a detailed list of hyperparameters.
To ensure a fair comparison for all models and ablations, we stick to hyperparameters presented in the table, except for the cases where we ablate certain parameters (\cref{tab:ablations}).
The only hyperparameters that change across benchmarks are the number of steps for the learning rate warmup and the number of input frames.
The number of warmup steps depends on the dataset size and corresponds to 10\% of the total number of steps, i.e.
\begin{equation}
    \# steps_{warmup} = 0.1 * \# steps/epoch * \#  epochs
\end{equation}

\begin{table}[tb]
  \caption{Hyperparameters for the base models for Charades-STA~\cite{gao2017charadesSTA}, QVHighlights (QVH)~\cite{lei2021qvhighlights}, and ActivityNet (ANet)~\cite{krishna2017ActivityNetCaptions} and for training of the answerer for NExT-GQA~\cite{Xiao2023NExT-QGA}. LR: Learning rate.}
  \label{tab:supp:Hyperparameters}
  \centering
  \begin{tabular}{@{}lc@{\hskip 0.5in}c@{}}
    \toprule
    \textbf{Hyperparameter} & \textbf{Charades-STA} & \textbf{QVH/ ANet/ NExT-GQA} \\
    \midrule
    Batch size & 32 & 32 \\
    Epochs  & 20 & 50 \\
    LR & 3e-4 & 3e-4 \\
    LR warmup & Linear & Linear \\
    LR warmup steps & 10\% & 10\% \\
    
    LR decay & Cosine & Cosine \\
    Optimizer & AdamW & AdamW \\
    Weight decay & 0.05 & 0.05 \\
    \# input frames & 20 & 60 \\
    \# beams & 5 & 5 \\
  \bottomrule
  \end{tabular}
\end{table}

\subsection{Zero-shot experiment details}
We query \textit{gpt-4o-2024-08-06} through the API.
We use \textit{low} image details for all queries.

We use temperature zero and maximum generation length for all the queries and use other API parameters as default.
We observe some non-determinism even when using temperature 0, therefore all GPT-4o results are averaged over at least two runs. 

\subsection{Video processing}

\label{sec:Setup_Implementation_Details_1}
We build on top of the Salesforce \href{https://github.com/salesforce/LAVIS/tree/main}{lavis} repository and leverage their implementation of randomly and uniformly extracting frames from a video.
During training, we randomly sample frames as a means of data augmentation to alter the frames seen by the model in each batch.
At inference, we sample uniformly to provide an equal coverage of the entire video.
For random sampling, we start by uniformly extracting n+1 timestamps, where the first and last timestamps are at $t=0$ and $t=video\_length$, respectively.
Afterward, we randomly sample one frame between each pair of adjacent timestamps, yielding the final n randomly sampled frames.
This process of random sampling can be considered adding noise to a uniform sampling process.

During training, we randomly crop and resize each frame to 224x224 pixels.
We then normalize each frame by subtracting by a fixed mean and dividing by a specified standard deviation.
At inference, we don't apply random cropping and resizing.
We only apply the normalization.

\section{Zero-shot Chrono Prompt for Zero-shot moment retrieval with GPT-4o}

In \cref{fig:Supp:GPT4o_prompt_1}, we show the prompt used for querying GPT-4o in the zero-shot setting.
This includes indications on the expected format, as well as encouragement to reason step by step before providing the final window.
\begin{figure*}[tb]
    \centering
    \begin{tcolorbox}[width=6in,
        boxrule=1.2pt,
        colback=white,
        arc=3mm]
    {\Large\textbf{User}}
    \vspace{2mm}

    [Frame at \textcolor{blue}{$t_1$} seconds: \textcolor{blue}{$f_1$}] [Frame at \textcolor{blue}{$t_2$} seconds: {\textcolor{blue}{$f_2$}}] [...] [Frame at \textcolor{blue}{$F$} seconds: \textcolor{blue}{$f_F$}] 
    \newline
    The video lasts {\textcolor{blue}{duration}} seconds.
    \newline
    Query: {\textcolor{blue}{query}}.
    \newline
    Given the video and the query, find the relevant windows.
    Think step by step.
    Reason about the events in the video and how they relate to the query. After your reasoning, output `ANSWER: $<$your answer$>$` in the format specified in the task prompt.
    Always provide a non-empty answer after your thoughts. If you think the event does not take place in the video, give your best guess, as otherwise the evaluation will be marked as incorrect.
    Never provide an empty list for $<$your answer$>$.
    The descriptions of moments are sometimes imprecise, so retrieve the closest moment.
    If you don't see an event remotely similar to the description, guess what is the most likely moment given the context.
    For instance, for cutting onion this could be between the time we see that the scene takes place in the kitchen and the time we see the onions being boiled in the pan.
    The answer should be in the format of a list indicating the start and end of a window of moment, [start\_window, end\_window], for instance [0, 1].
    If you detect multiple windows for the same moment, choose the most relevant one.
    It's important your final answer only contains one window. It is very important that the answer is in this format, otherwise the evaluation will fail.

    \vspace{5mm}
    \hrule
    \vspace{5mm}
    {\Large\textbf{Assistant}}
    \vspace{2mm}
    
    \textcolor{blue}{Answer}
    \end{tcolorbox}
    \caption{Prompt used for GPT-4o for single-window moment localization. \textcolor{blue}{Blue} text represents variables.}
    \label{fig:Supp:GPT4o_prompt_1}
\end{figure*}

\section{Additional Qualitative Results}
\label{sec:Supp:Additional_Qualitative_Results}

In this section, we provide further qualitative results and discuss shortcomings of the proposed approach which can be addressed in future research.
For each example, we provide the discussion in the respective caption for the convenience of the reader. 
Examples 1 through 6 are from QVHighlights~\cite{lei2021qvhighlights} and examples 6 through 10 are from Charades-STA~\cite{gao2017charadesSTA}.

As in the main paper, we illustrate the ground truth targets as dashed lines and the predicted windows as solid lines.

\begin{figure*}[tb]
  \centering
  \includegraphics[width=\textwidth]{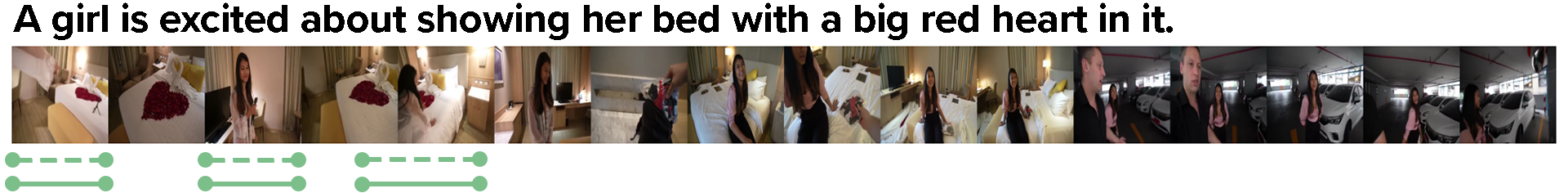}
  \caption{
  We observe the capability of the model using the \chronoWatch blueprint to recognize and differentiate 3 separate moments that repeatedly occur with intermittent interruptions.
  Each predicted moment depicts the respective natural language query.
  }
  \label{fig:Appendix-QVH_1}
  
\end{figure*}

\begin{figure*}[tb]
  \centering
  \includegraphics[width=\textwidth]{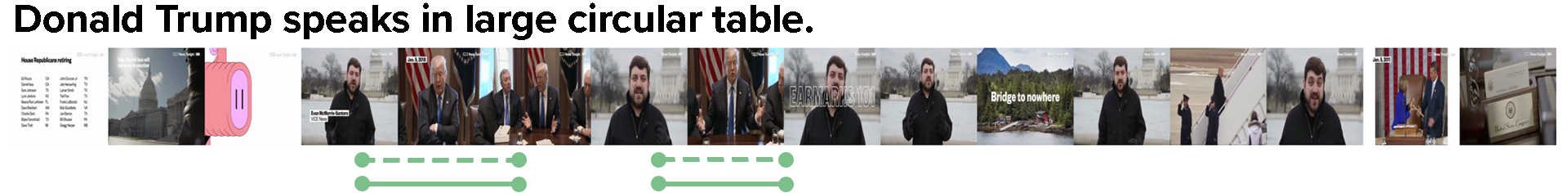}
  \caption{
    \chronoBlip can recognize a distinct public figure, Donald Trump, although the training set includes only 6 queries containing Donald Trump.
  }
  \label{fig:Appendix-QVH_2}
  
\end{figure*}
\begin{figure*}[tb]
  \centering
  \includegraphics[width=\textwidth]{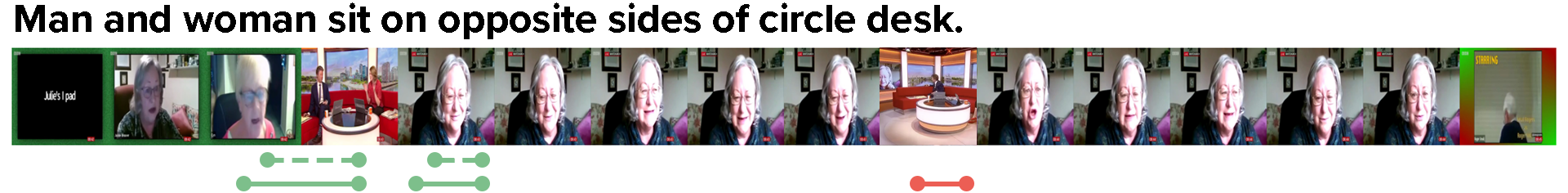}
  \caption{
    The model using the \chronoWatch blueprint predicts a third window, that does not align with a ground truth moment, which, in fact, is accurate and depicts the query.
  }
  \label{fig:Appendix-QVH_3}
  
\end{figure*}

\begin{figure*}[tb]
  \centering
  \includegraphics[width=\textwidth]{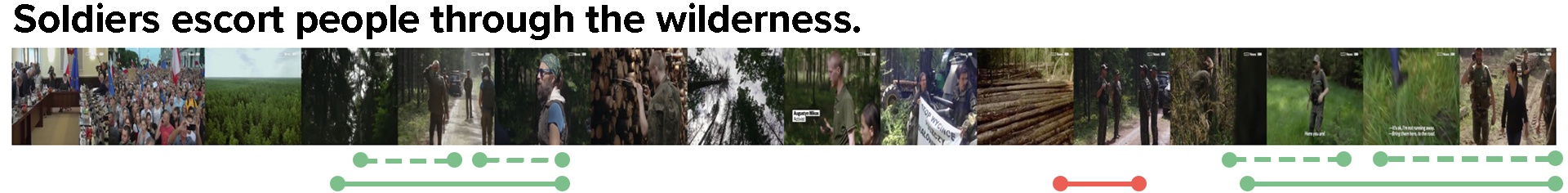}
  \caption{
    The video contains two groups of relevant moments which contain a short sub-2-second cut that does not depict the query.
    With a resolution of 60 frames for a 150-second long video, corresponding to a frame being seen every 2.5 seconds, \chronoBlip can not detect the cut and predicts two long moments that encompass the two short ones.
    Moreover, the model using the \chronoWatch blueprint predicts a third window (red) that does not match a ground truth label.
    The predicted moment does in actuality depict soldiers, but those are not escorting other people.
  }
  \label{fig:Appendix-QVH_5}
  
\end{figure*}

\begin{figure*}[tb]
  \centering
  \includegraphics[width=\textwidth]{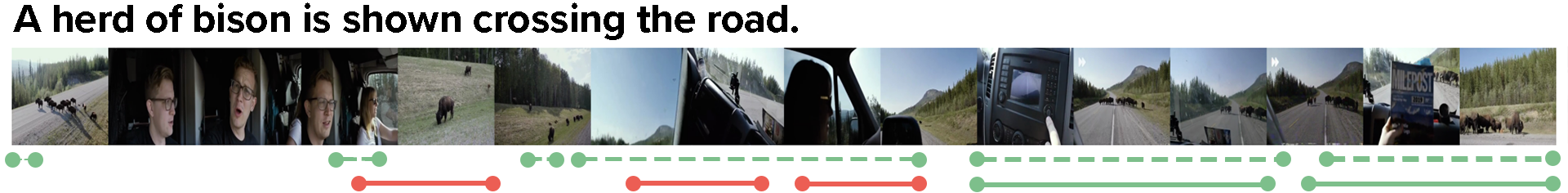}
  \caption{
    The model using the \chronoWatch blueprint has trouble predicting very short moments in long videos and high-frequency jump cuts given the resolution of the frame sampling.
  }
  \label{fig:Appendix-QVH_6}
  
\end{figure*}

\begin{figure*}[tb]
  \centering
  \includegraphics[width=\textwidth]{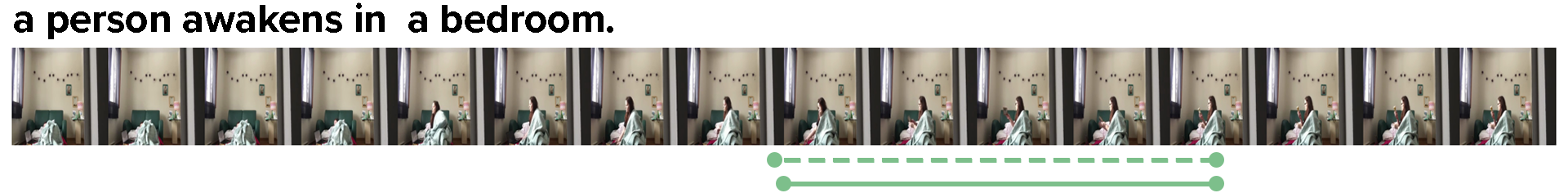}
  \caption{
  Given a very ambiguous query, the prediction of the model using the \chronoWatch blueprint aligns with the ground truth.
  }
  \label{fig:Appendix-Charades_3}
  
\end{figure*}

\begin{figure*}[tb]
  \centering
  \includegraphics[width=\textwidth]{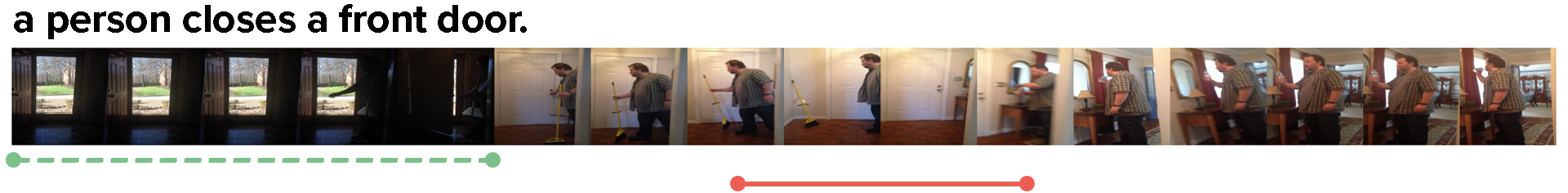}
  \caption{
    The model using the \chronoWatch blueprint fails to recognize the action of the man closing the front door.
    We hypothesize this is because the door is very dark and poorly visible.
    The model using the \chronoWatch blueprint predicts the moment when the door is clearly visible and closed.
  }
  \label{fig:Appendix-Charades_5}
  
\end{figure*}

\begin{figure*}[tb]
  \centering
  \includegraphics[width=\textwidth]{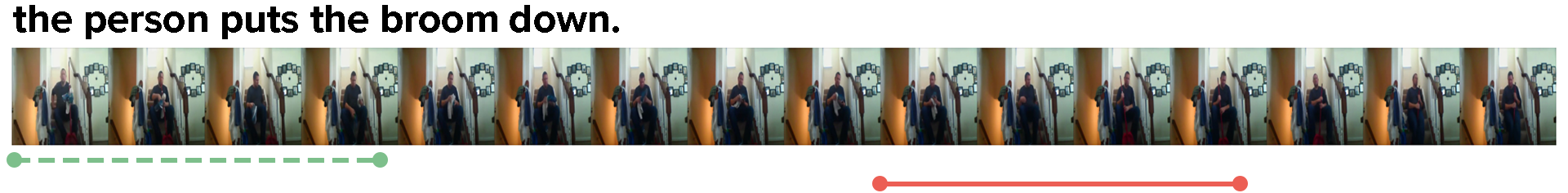}
  \caption{
  The model using the \chronoWatch blueprint fails to recognize the action of putting down the broom at the beginning of the video and predicts the moment when the man picks the broom back up and then holds it.
  }
  \label{fig:Appendix-Charades_6}
  
\end{figure*}

\begin{figure*}[!t] 
  \centering
  \includegraphics[width=\textwidth]{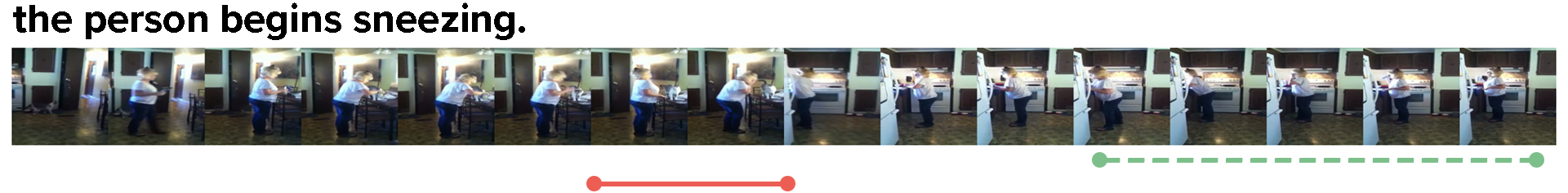}
  \caption{
      \chronoBlip recognizes the moment when the person is sneezing.
      The ground truth is false.
  }
  \label{fig:Appendix-Charades_7}
  
\end{figure*}

\end{document}